\documentclass{article}

\usepackage{microtype}
\usepackage{graphicx}
\usepackage{subfigure}
\usepackage{booktabs} 
\usepackage{textcomp}

\usepackage{hyperref}

\usepackage{enumitem}

\usepackage[accepted]{icml2024}

\usepackage{amsmath}
\usepackage{amssymb}
\usepackage{mathtools}
\usepackage{amsthm}
\usepackage{multirow}
\usepackage[capitalize,noabbrev]{cleveref}
\theoremstyle{plain}

\theoremstyle{definition}

\theoremstyle{remark}

\usepackage[textsize=tiny]{todonotes}

\icmltitlerunning{VoroNav: Voronoi-based Zero-shot Object Navigation with Large Language Model}

\begin{document}

\twocolumn[
\icmltitle{VoroNav: Voronoi-based Zero-shot Object Navigation \\
with Large Language Model}



\icmlsetsymbol{equal}{*}

\begin{icmlauthorlist}
\icmlauthor{Pengying Wu}{yyy,equal}
\icmlauthor{Yao Mu}{sch,comp,equal}
\icmlauthor{Bingxian Wu}{yyy}
\icmlauthor{Yi Hou}{yyy}
\icmlauthor{Ji Ma}{yyy}
\icmlauthor{Shanghang Zhang}{yyy}
\icmlauthor{Chang Liu}{yyy}

\end{icmlauthorlist}

\icmlaffiliation{yyy}{Peking University}
\icmlaffiliation{comp}{OpenGVLab, Shanghai AI Laboratory}
\icmlaffiliation{sch}{The University of Hong Kong}

\icmlcorrespondingauthor{Shanghang Zhang}{shanghang@pku.edu.cn}
\icmlcorrespondingauthor{Chang Liu}{changliucoe@pku.edu.cn}


\vskip 0.3in
]



\printAffiliationsAndNotice{\icmlEqualContribution} 

\begin{abstract}
In the realm of household robotics, the Zero-Shot Object Navigation (ZSON) task empowers agents to adeptly traverse unfamiliar environments and locate objects from novel categories without prior explicit training. This paper introduces VoroNav, a novel semantic exploration framework that proposes the Reduced Voronoi Graph to extract exploratory paths and planning nodes from a semantic map constructed in real time. 
By harnessing topological and semantic information, VoroNav designs text-based descriptions of paths and images that are readily interpretable by a large language model (LLM).
In particular, our approach presents a synergy of path and farsight descriptions to represent the environmental context, enabling LLM to apply commonsense reasoning to ascertain waypoints for navigation. 
Extensive evaluation on HM3D and HSSD validates VoroNav surpasses existing benchmarks in both success rate and exploration efficiency (absolute improvement: +2.8\% Success and +3.7\% SPL on HM3D, +2.6\% Success and +3.8\% SPL on HSSD). 
Additionally introduced metrics that evaluate obstacle avoidance proficiency and perceptual efficiency further corroborate the enhancements achieved by our method in ZSON planning. 
\textbf{Project page: }\href{https://voro-nav.github.io}{https://voro-nav.github.io}

\end{abstract}

\label{sec:intro}

\section{Introduction}

Navigation capability holds great significance for household robots, enabling these machines to effectively reach designated areas and complete various subsequent tasks. 
Within this context, Zero-Shot Object Navigation (ZSON) demands 
\begin{figure}[H]
  \centering
   \includegraphics[width=0.8\linewidth]{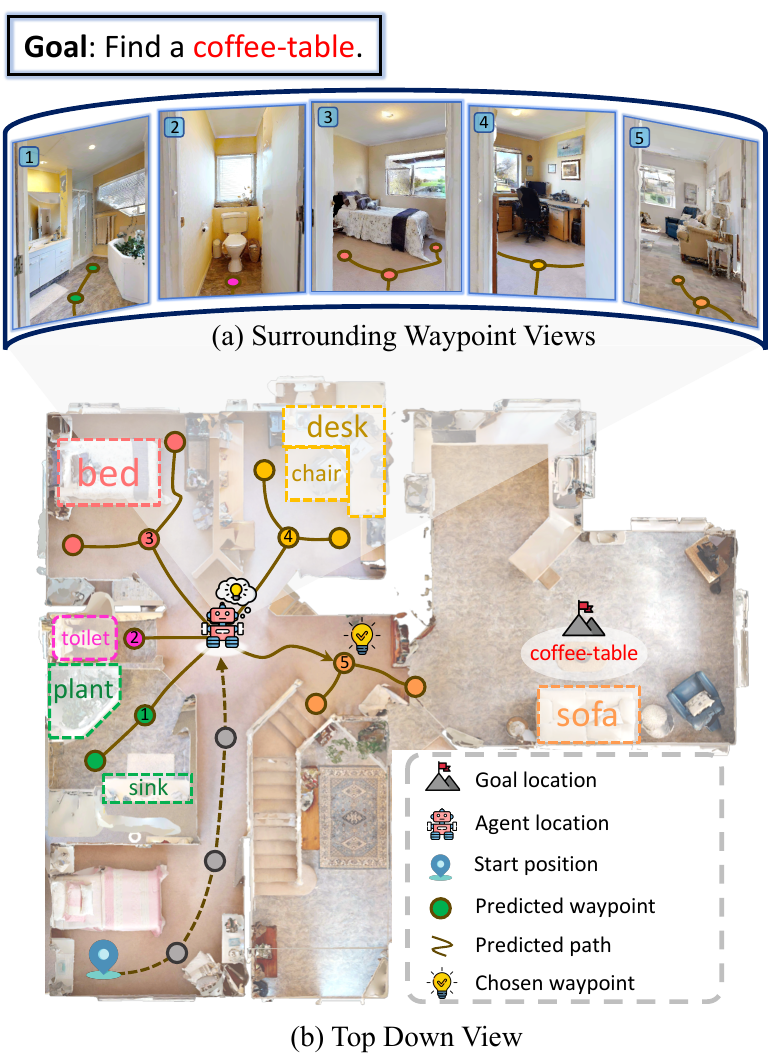}

   \caption{
   \textbf{Voronoi-based Navigation with LLM.} Our model focuses on optimizing the decision-making process in ZSON. 
   It enables the agent to pinpoint intersections rich in observation on the map by Voronoi sparsification, which act as navigation waypoints. 
   The agent perceives the environment at intersections, collects scene information from nearby waypoints, and performs reasoning guided by LLM to ascertain the most plausible waypoint leading to the desired target. 
   The five images presented in (a) depict the agent's corresponding perspectives as it faces five adjacent navigation waypoints at the intersection illustrated in (b), with the indices showing the correspondence.
   }
   \label{fig:head}
\end{figure}
that an agent have the ability to move toward a target object of an unfamiliar category by leveraging scene reasoning, a capability essential for the performance of diverse complex tasks by household robots. 
The core of ZSON centers on making good use of general commonsense to steer agents for exploration with minimal movement cost and accurate localization of a novel target object. 

Current ZSON methods can be categorized into two types: end-to-end, network-based navigation \cite{majumdar2022zson, 10161345, 10161289, chen2023zeroshot, gadre2023cows}; and modular, map-based navigation \cite{zhou2023esc,gadre2023cows,chen2023train}. 
The end-to-end methods use reinforcement or imitation learning for training policy networks, and are designed for mapping directly from RGB-D images to actions. However, the end-to-end model's output lacks interpretability and necessitates a substantial amount of training data, and exhibits serious inefficiency problems of back-and-forth redundant movement regarding the actual performance. 

Map-based methods leverage maps to store historical topology information for planning purposes. 
Map-based navigation frameworks usually plan new waypoints either every predetermined number of steps or when the increment in map building reaches a specific threshold. 
However, the selected waypoints 
usually come short of optimal positions for decision-making. 
This occurs because the agent could arrive at an intersection with massive information and potentially uncover expansive unseen areas by just one more step from here, which can bring huge benefits to scene reasoning and task planning. 
Yet, such benefits may not be obtained using the traditional strategy of choosing waypoints, for they would not actively identify informative points as waypoints. 
Just imagine that you are looking for an object, walking down a long corridor, and encountering the scene shown in \cref{fig:head} (b). Would you be more inclined to halt at the intersection, take a moment to observe your surroundings, and then make a thoughtful decision after comparing the adjacent areas? 
Therefore, this study puts insight into the positive impact of making decisions at intersections in the field of navigation, and develops a Reduced Voronoi Graph (RVG) generation approach to distill intersection points and viable pathways from the real-time built map. 
Utilizing graph-structured RVG, we systematize the planning process as navigation subtasks across graph nodes.

Another significant issue faced by existing navigation algorithms is the integral representation of observed scenes for subsequent planning. When presented with RGB images, network-based approaches leverage semantic embeddings to identify novel object categories and utilize recurrent policy networks to directly predict optimal actions \cite{khandelwal2022simple}; 
Conversely, map-based methods mostly employ an open-set detector to segment RGB images, which, in conjunction with depth data and pose information, are utilized to construct a semantic map. By interpreting the representation of the semantic map, the next subgoal point is selected \cite{chaplot2020object}. 
Each method, however, presents distinct limitations: network-based methods struggle with low exploration efficiency and constrained planning memory that is limited by implicit scene representation and network size, whereas map-based methods only build maps within the field of view of the depth camera, thus unable to integrate information beyond the depth sensing range to plan informed waypoints. 
To overcome these limitations, it is essential for the agent to fuse the observations of both maps and images, comprehensively understand both modalities, and make appropriate decisions. 

To provide a direct response, we adopt the large language model (LLM) as a cognitive engine for spatial reasoning to understand various scenes. The custom-designed prompts are developed to effectively integrate observations of maps and images, considering the preference of LLM. Previous studies \cite{zhou2023esc, yu2023l3mvn} have collected objects around frontier points to depict the scenes of the exploration areas, and then employed LLMs to infer probable locations of the target. 
While these foundational applications established a groundwork for guiding navigation with LLM, we reimagine these principles by imitating
human exploratory behaviors. 
Specifically, human exploration typically involves scene description from two perspectives: the egocentric view and the scenes along traversable paths. 
Descriptions that align analogously with human cognition ensure that the resulting prompts are closer to human corpora \cite{beckner2009language, lai2018language}, and previous works \cite{NEURIPS2020_1457c0d6, naveed2023comprehensive} show that LLM typically exhibits enhanced performance when dealing with natural language problems similar to the corpora. Starting from this standpoint, we formulate the prompt by generating descriptions of paths (scenes along traversable paths) and farsight images (egocentric view), thereby promoting LLM's understanding of the observed scenes.

The ZSON task requires the agent to find the target at the lowest path cost. The reasoning results of LLM can guide the agent in predicting the probable locations of the target, but struggle to handle the problem of exploration. So we design a hierarchical reward mechanism that combines the topological information of the map and the suggestions provided by LLM. This mechanism evaluates the exploration significance, path efficiency, and commonsense tendencies of each candidate waypoint on the RVG, thereby facilitating more informed decision-making.

Our contributions can be summarized as follows:
\vspace{-10pt}
\begin{itemize}[leftmargin=10pt, itemsep=-2pt]
\item We introduce Voronoi-based scene graph generation for ZSON, designed to select waypoints that provide a wealth of observation data to facilitate subsequent planning processes.

\item We design an innovative prompting strategy of scene representation that combines both path and farsight descriptions to provide holistic scene descriptions for LLM to analyze and evaluate.

\item We propose a decision-making policy that necessitates deliberation among exploration, path efficiency, and commonsense tendencies to yield rational actions. 

\item We achieve state-of-the-art results on the ZSON task and outperform benchmark methods on representative datasets, i.e., HM3D \cite{ramakrishnan2021habitat} and HSSD \cite{khanna2023habitat}.

\end{itemize}

\begin{figure*}[tbp]
  \centering
   \includegraphics[width=0.85\linewidth]{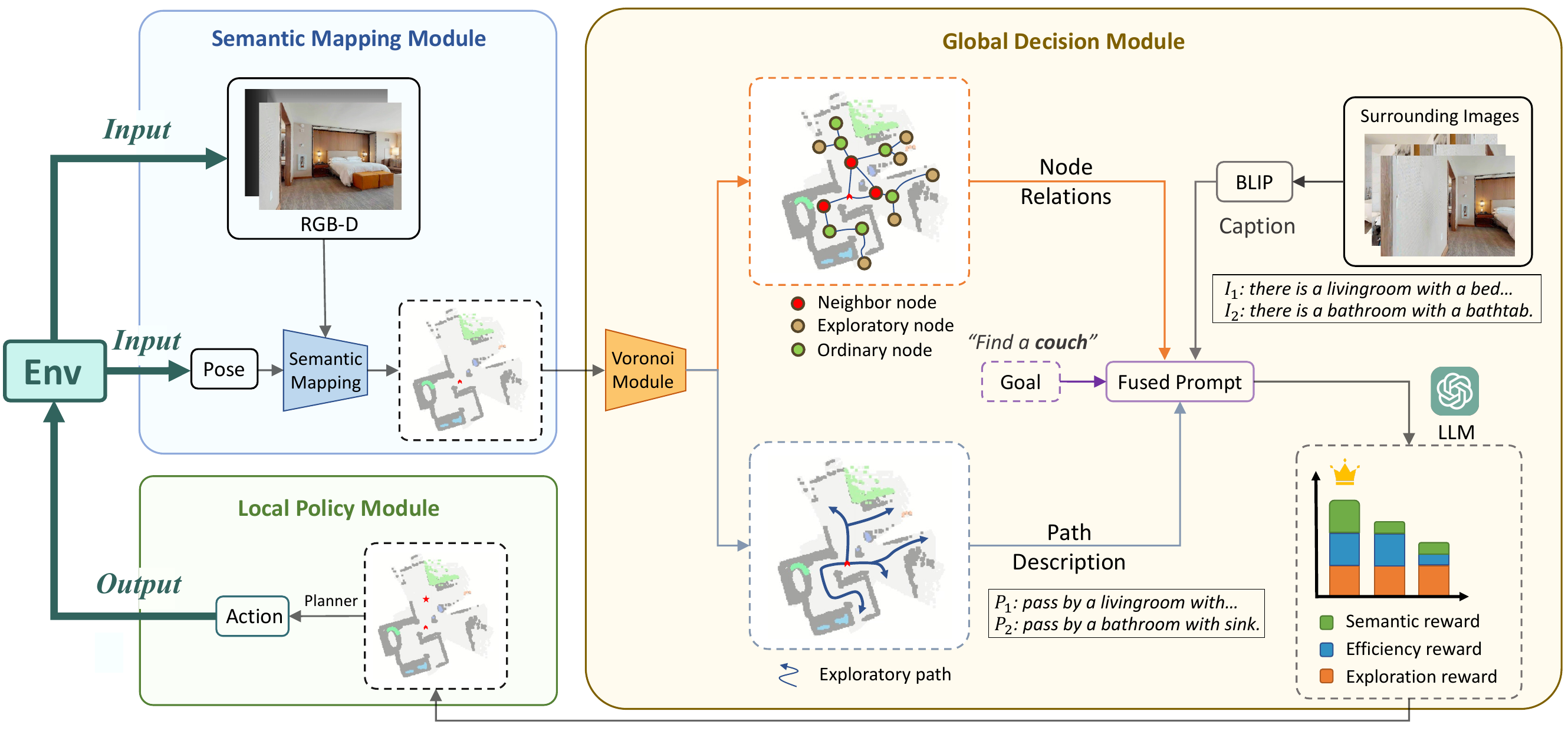}

   \caption{\textbf{Components of VoroNav}. VoroNav includes three modules. Perceptual inputs include RGB-D images and real-time pose, while the output of the agent is ``Action''. The RGB-D and pose observation are processed by the Semantic Mapping Module (light blue module) to form a semantic map. The Global Decision Module (light yellow module) generates RVG, which is used to produce textual descriptions of surrounding neighbor nodes and exploratory paths. This module then employs an LLM to assist in selecting the promising neighbor node as a mid-term goal by inferring the fused prompt of scene descriptions. The Local Policy Module (light green module) plans the low-level actions of the agent to reach the target point.}
   \label{fig:model}
   \vspace{-3pt}
\end{figure*}

\section{Related Work}
\label{sec:rel}

\subsection{Zero-shot Object Navigation}
In contrast to conventional object navigation, ZSON aims to locate objects of unfamiliar categories and attain high exploration efficiency. 
Image-based ZSON works \cite{majumdar2022zson,al2022zero, gadre2023cows} map the egocentric images and target object instructions to the embedding spaces, utilizing a trained policy network to predict subsequent actions. 
In contrast, map-based ZSON works mostly adopt hierarchical structures, integrated with zero-shot object detectors that identify target objects. These approaches make informed decisions by leveraging prior knowledge of object relationships \cite{chen2023train} or by employing large language models \cite{zhou2023esc,yu2023l3mvn,shah2023lfg}.

\subsection{Scene Representation for Navigation}
In the hierarchical framework of visual navigation, scene representation is used to process the received observation information into an explicit structure that can be directly utilized by subsequent decision-making. 
Frontier-based works \cite{ramakrishnan2022poni,chen2023train,gadre2023cows,gervet2023navigating}
model semantic information into frontiers extracted by online grid maps to complete exploration with specific tasks. 
Graph-based works predict waypoints directly from RGB-D images \cite{krantz2020beyond, krantz2021waypoint,an2023etpnav} or from sparsified maps \cite{li2020improving,kwon2023renderable,liu2023revolt} to represent the environment as topological maps, integrating geographic and semantic information into nodes for waypoint navigation. 

\subsection{LLM Guided Navigation}
LLMs have become a new way of prior-knowledge reasoning in navigation due to its powerful information processing and generative capabilities. 
For example, \citet{zhou2023esc} use LLM to predict the degree of correlation with the target object at the object level and the room level to infer where is the most likely location of the target object. 
\citet{yu2023l3mvn} generate clusters of unexplored areas by frontiers, and leverage LLM to infer the correlation between the target object and the objects contained within each cluster to navigate to the scene closer to the target object. 
\citet{gadre2023cows} adopt LLM to provide prior information at the object level to assist in target object localization. 
\citet{shah2023lfg} feed chain-of-thought(CoT) into LLM for navigation that encourages exploration of areas with higher relevance while concurrently avoiding moving to areas that are unrelated to the target object. 
\citet{cai2023bridging} cluster panoramic images into scene nodes by LLM, use CoT of LLM to determine whether exploration or exploitation, select the image with the highest likelihood of finding the target object, and navigate accordingly based on the chosen image. 
\citet{yu2023conavgpt} apply the decision-making of LLM for multi-robot collaborative navigation, and the LLM centrally plans the mid-term goal for each robot by extracting information such as obstacles, frontiers, object coordinates, and robot states from online maps.

\section{VoroNav Approach}
\label{sec:meth}
This section first introduces the task definition of ZSON (\Cref{sec:task_def}). 
Subsequently, the modules of the VoroNav framework are introduced. 
As shown in \cref{fig:model}, VoroNav constructs a semantic map in Semantic Mapping Module (\Cref{sec:mapping}), then determines the mid-term goal in Global Decision Module (\Cref{sec:global}), and finally plans local motion and selects a discrete action in Local Policy Module (\Cref{sec:local}).

\subsection{Task Definition of ZSON}
\label{sec:task_def}
Traditional supervised object navigation relies on the knowledge or reward from the training data to predict the optimal action $a_t$ and is limited to navigating to targets within a closed set of known categories $\mathcal{K}$. However, the ZSON task requires neither purposeful training nor closely linked prior knowledge for navigation toward a novel set of object types $\mathcal{N}$.
Initially, the agent is placed at a designated start point $p_0$ and is given the category $G \in \mathcal{N}$ of the target to find. 
The agent's observation includes RGB-D images $I_t$ and the real-time pose $p_t$ in the environment $\mathcal{E}$. 
An effective decision-making framework needs to be developed to leverage these observed data $\mathcal{O}_t=\{${\small $\{p_0, I_0\}, \ldots, \{p_t, I_t\}$}$\}$ to understand and deduce the environment, aiming to predict the likely position of the target object. 
The agent is required to explore the environment according to its planning module until it discovers the target, after which it should proceed toward the target.
Success is achieved when the agent reaches a geodesic distance of less than $0.1$ meters from the target and executes a ``Stop'' command. 
Conversely, the task is deemed failed if the agent either exceeds the maximum step count without finding the target or executes the ``Stop'' action at a distance greater than $0.1$ meters from the target.

\subsection{The Semantic Mapping Module}
\label{sec:mapping}
We maintain a 2D semantic map $\mathcal{M}_t$ by processing RGB-D images $\{I_0, \ldots, I_t\}$ and poses $\{p_0, \ldots, p_t\}$. This semantic map is structured as a $(K+2)\times M \times M $ grid, where $M$ denotes the dimensions of the map's width and height, and $(K+2)$ indicates the total number of channels within the map. 
These channels comprise $K$ categorical maps, an obstacle map, and an explored map, which correspond to detected object regions, obstacle regions, and observed regions, respectively. 
Given the depth image and the agent's pose, 3D point clouds are generated. All point clouds near the floor are assigned to the explored map representing the feasible area to travel through, whereas those at other heights are mapped into the obstacle map. Meanwhile, we predict the category masks of the RGB image by Grounded-SAM \cite{liu2023grounding,kirillov2023segany} and map the masks into 3D semantic point clouds using the depth information and the agent's pose. The 3D point clouds with $K$ categorical information are correspondingly mapped to $K$ categorical map channels.

\subsection{The Global Decision Module}
\label{sec:global}
{\bf Graph Extraction.}
The Generalized Voronoi Diagram (GVD) of a map depicts a set of points that are equidistant from the two closest obstacle points, representing the medial-axis pathway of unoccupied space outside the obstacles of arbitrary shape. Let $\mathcal{X} \in 
\mathbb{R}^{2}$ be the map space and $\Omega$ denote the area occupied by obstacles on the map. The point set $\mathcal{V}$ of GVD can be represented as follows: 

\vspace{-20pt}
\begin{equation}
  \small \mathcal{V}=\{x \in \mathcal{X} \backslash \Omega |\exists \omega_i \neq \omega_j \in \Omega, d (x,\omega_i)= d (x,\omega_j)=f(x)\}
  \label{eq:voronoi}
\end{equation}
where $\omega_{(\cdot)}$ represents any point within the obstacles $\Omega$, the function $d(\cdot, \cdot)$ denotes the Euclidean distance between two points, while $f(\cdot)$ signifies the positive \textit{Euclidean Signed Distance Field} (ESDF), which is defined as follows:

\vspace{-15pt}
\begin{equation}
  f(x)=\inf _{y \in \partial \Omega} d (x, y)
  \label{eq:esdf}
\end{equation}
where $\partial\Omega$ indicates the boundary of the obstacles.

Given the obstacle and explored maps, we can obtain the GVD points from these maps and construct RVG $\mathcal{G}$ to represent the observed spaces\footnote{The process for generating RVG is detailed in \cref{appendix_RVG} }.
We then classify the RVG nodes into four categories based on the node positions: \textit{agent nodes}, \textit{neighbor nodes}, \textit{exploratory nodes}, and \textit{ordinary nodes}. 
Specifically, the node closest to the agent is designated as the agent node, representing the agent's current decision-making position; 
The nodes directly connected to the agent node are considered neighbor nodes for subsequent planning; 
The nodes adjacent to unexplored areas with a single connecting edge are classified as exploratory nodes. 
All the other nodes are categorized as ordinary nodes.

\begin{figure}[htbp]
  \centering
   \includegraphics[width=0.85\linewidth]{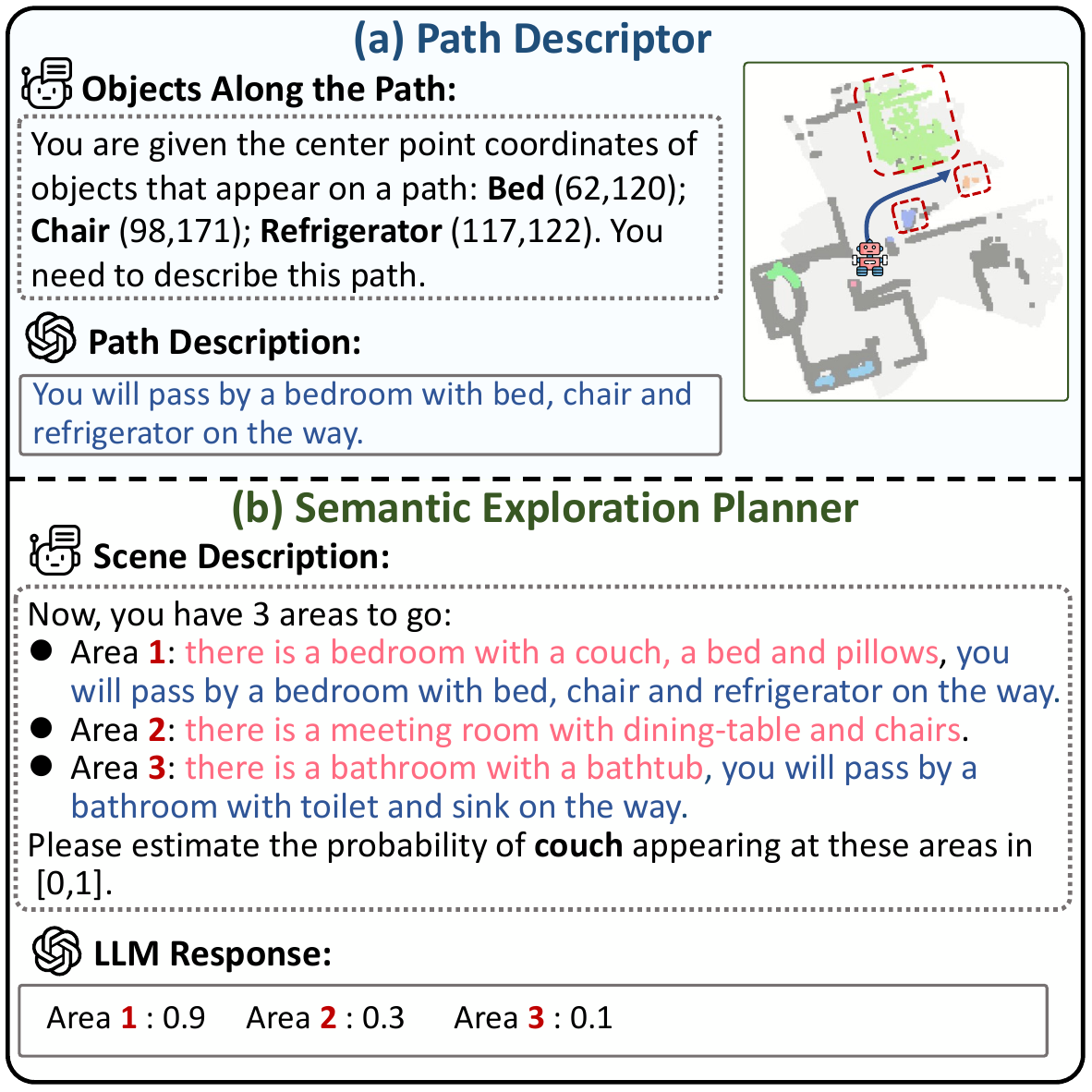}

   \caption{\textbf{Commonsense Reasoning with LLM.} (a) LLM analyzes the objects and their coordinates that appear on the path and depicts the scene along the path. (b) LLM predicts the probability of the target object appearing in each area by comprehending the fused text descriptions of the scene.
   }
   \label{fig:conversation}
   \vspace{-18pt}
\end{figure}

{\bf Path Description.}
We generate navigable paths formed by RVG edges and create text descriptions that embody the scene along each path, as shown in \Cref{fig:conversation} (a).
To be specific, given $m$ exploratory nodes, we leverage the Wavefront Propagation method \cite{kalra2009incremental} to obtain the shortest path $P_j$ from the agent node to the $j^{\text{th}}$ exploratory node on the GVD and compile all the paths into a set $\mathcal{P}=\{P_1, \ldots, P_m\}$, as shown in \Cref{fig:voronoi} (c). 
To generate the semantic description of each path $P_j$, we gather from the semantic map the occurrence of $c$ objects $\{o_{j,1}, \ldots, o_{j, c}\}$ along the exploratory path $P_j$ and the objects' central locations $\{l_{j, 1}, \ldots, l_{j, c}\}$. Assuming paths within the set $\{P_a, \ldots, P_b\}$ all pass through the neighbor node $N_i$, prompts of the form Template$(\{${\small $P_a:(o_{a, 1}, l_{a, 1}), \ldots$}$\} \bigcup \ldots \bigcup \{${\small $P_b:(o_{b, 1}, l_{b, 1}), \ldots$} $\})$ are generated for the neighbor node $N_i$, by collecting and summarizing the semantic information along the paths in $\{P_a, \ldots, P_b\}$. 
The function Template$(\cdot)$ processes the input data, converting it into the textual form and integrating it with predefined templates to create a format conducive to conversational interactions with LLM (refer to \Cref{fig:conversation} (a): \textit{Objects Along the Path}).
Afterward, to distill the fragmented and unstructured path information into a coherent format, we employ GPT-3.5 \cite{ouyang2022training}, which possesses robust comprehension and generative capabilities, for creating the scene descriptions $D^p_i$ along paths that traverse each neighbor node $N_i$ (refer to \Cref{fig:conversation} (a): \textit{Path Description}). 
Similarly, assuming there involve $n$ neighbor nodes, we describe the scenes of the paths each neighbor node $N_i$ leads to and compile the path descriptions into a set $\mathcal{D}^p = \{D^p_1, \ldots, D^p_n\}$. 
This process textualizes the scenarios the agent will encounter along possible navigable paths after reaching each neighbor node.

\begin{figure}[t]
  \centering
   \includegraphics[width=0.8\linewidth]{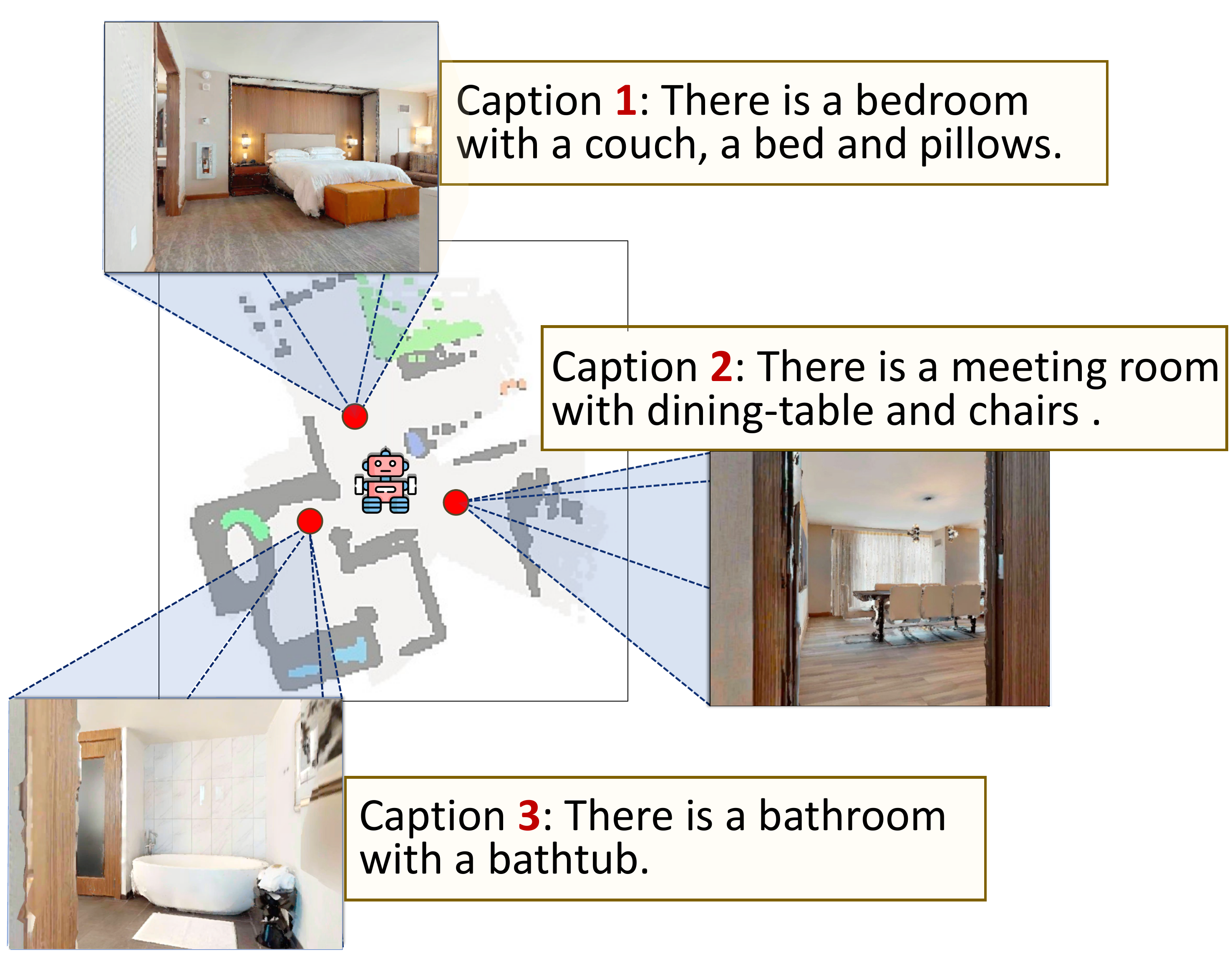}

   \caption{\textbf{Farsight Image Captioning.}
   The agent selects all RGB images that capture the views of neighbor nodes and uses BLIP to generate captions of these images.  
   }
   \label{fig:caption}
   \vspace{-10pt}
\end{figure}

{\bf Farsight Description.}
Path description generation is the process of converting the semantic map into scene descriptions of path form; however, 
the semantic map is constrained by the depth camera's limited range, precluding the incorporation of map information beyond its scope. 
Consequently, semantic descriptions of RGB images of unexplored areas add crucial complementary context for robot navigation.
As shown in \cref{fig:caption}, at the onset of the ZSON task or upon reaching an RVG node, the agent executes a full rotation to capture panoramic images. 
We then determine the ray $R_i$ on the map that extends from the agent's current node (agent node) to each neighbor node $N_i$. The RGB image $I_k$ collected from the full rotation $\mathcal{I}_t=\{I_{t-11}, \ldots, I_{t}\}$ (a full rotation includes $12$ turns), whose central \textit{Line of Sight} (LoS) $T_k$ exhibits the least angular deviation from the ray $R_i$, is identified as the one ($ I^f_i$) oriented towards the corresponding neighbor node $N_i$. Let $\mathcal{T}_t=\{T_{t-11}, \ldots, T_{t}\}$ be the central LoS set of $\mathcal{I}_t$, the process of matching images with each neighbor node $N_i$ can be defined as follows:

\vspace{-15pt}
\begin{equation}
\begin{array}{cl}
\mathop{\arg \min}\limits_{T_k} & 
g(R_i, T_k)\\
\text { s.t. } & T_k \in \mathcal{T}_{t}
\end{array}
  \label{eq:view}
\end{equation}
\vspace{-15pt}

where the function $g(\cdot, \cdot)$ indicates the angle between two rays on the map.
The BLIP model \cite{li2022blip} is then employed to generate descriptions $\mathcal{D}^f = \{D^f_1, \ldots, D^f_n \}$ for those images $\{I^f_1, \ldots, I^f_n \}$ facing different neighbor nodes $\{N_1, \ldots, N_n \}$.

{\bf Planning with LLM.} 
We select the mid-term target points by considering three distinct factors: exploration objective, locomotion efficiency (traversed path length), and alignment with typical scene layouts. 
The rewards for exploration and efficiency are space reasoning results stemming from spatial topology. To encourage the agent to explore the environment, we design a binary exploration reward vector $\mathbf{P}$ to indicate if there exists an exploratory path from the agent node to each exploratory node that traverses through neighbor nodes. Considering exploration efficiency, we design an efficiency reward vector $\mathbf{C}$ and assess whether each neighbor node lies within the previously traversed area.

Semantic rewards are reasoning feedback grounded in empirical knowledge and commonsense. The Global Decision Module utilizes the commonsense reasoning capabilities of the large language model, GPT-3.5, to select the most promising goal node for finding or approaching the target object among all neighbor nodes.
To this end, we combine the path and farsight descriptions of each neighbor node to generate a specially designed form of prompt that is amenable for GPT-3.5. This enables the LLM to more accurately estimate the probability of the target object's presence on each neighbor node by detailed prompts, as illustrated in \cref{fig:conversation} (b).
The probabilities given by the LLM's response are compiled into a semantic reward vector $\mathbf{L}$ and serve as varying levels of semantic incentive to navigate towards neighbor nodes.

When the agent simultaneously considers exploration, efficiency, and semantic aspects of decision-making, balancing the priority among these factors becomes challenging. To mitigate potential conflicts, we have implemented a hierarchical structure within the reward system. Assuming there are a total of $n$ neighbor nodes, the cumulative reward vector $\mathbf{W} \in \mathbb{R}^{n}$ is the sum of exploration reward vector $\mathbf{P}$, efficiency reward vector $\mathbf{C}$ and semantic reward vector $\mathbf{L}$.
The next navigation point selection can be formulated as follows:

\vspace{-10pt}
\begin{equation}
\begin{array}{cl}
\mathop{\arg \max}\limits_{\mathbf{S}} & \mathbf{W}^\text{T}\mathbf{S} \\
\text { s.t. } & \mathbf{S} \in \mathbb{E}^{n}
\end{array}
  \label{eq:planning}
\end{equation}
\vspace{-10pt}

where $\mathbb{E}^{n}=\{\mathbf{e}_1, \mathbf{e}_2, \ldots, \mathbf{e}_n\}$ is the standard orthogonal basis composed of n-dimensional coordinate vectors. 
The decision variable $\mathbf{S}=\mathbf{e}_i$ if $i^\text{th}$ neighbor node is selected for next navigation waypoint.
The reward vectors are defined as follows: 

\vspace{-10pt}
\begin{equation}
\left\{\begin{array}{l}
\mathbf{W}=\mathbf{P}+\mathbf{C}+\mathbf{L}, \\
\mathbf{P}=2\left(\alpha_{1} \mathbf{e}_{1}+\ldots+\alpha_{n} \mathbf{e}_{n}\right), \\
\mathbf{C}=\beta_{1} \mathbf{e}_{1}+\ldots+\beta_{n} \mathbf{e}_{n}, \\
\mathbf{0} \le  \mathbf{L} \le  \mathbf{1}, \\
\alpha_{i}, \beta_{i} \in\{0,1\}, \\
\mathbf{W}, \mathbf{P}, \mathbf{C}, \mathbf{L} \in \mathbb{R}^{n} .
\end{array}\right.
  \label{eq:defination}
\end{equation}
\vspace{-10pt}

where the $i^\text{th}$ dimensional component of $\mathbf{L}$ is the semantic score of the $i^\text{th}$ neighbor node provided by LLM within the interval $(0, 1)$. 
The binary coefficient $\alpha_i \in \{0, 1\}$ stands for whether the $i^\text{th}$ neighbor node is traversed through by exploratory paths ($\alpha_i=1$) or not ($\alpha_i=0$), and $\beta_i \in \{0, 1\}$ denotes whether the $i^\text{th}$ neighbor node is covered by historical trajectories ($\beta_i=0$) or not ($\beta_i=1$). We establish the hierarchy of priorities for each aspect by assigning different reward weights of the reward vectors as shown in \Cref{eq:defination} ($1^{\text{st}}:$ Exploration; $2^{\text{nd}}:$ Efficiency; $3^{\text{rd}}:$ Semantic).

We select the neighbor node that offers the highest cumulative reward as the next target waypoint for navigation. If the agent's vision model identifies the target object $G$ while exploring, the Semantic Mapping module will map the target's point cloud onto the existing map $\mathcal{M}$, enabling direct path planning toward the target's location.

\subsection{The Local Policy Module}
\label{sec:local}
Given the agent's pose, obstacle map, and target point, we use the Fast Marching Method \cite{sethian1996fast} to find the shortest path from the current position to the target, which is composed of a sequence of discrete points in the map. The nearest coordinate on this shortest path is selected as an immediate navigation objective for executing actions such as moving forward or turning. Once arriving at a Voronoi node, the agent will rotate and repeat the selection of the mid-term goal.

\section{Experiments}
\label{sec:exp}

\begin{table*}[htbp]
  \caption{\textbf{Comprison with ZSON Baselines.}
  Our proposed VoroNav outperforms the ZSON baselines on both HM3D and HSSD. To guarantee the zero-shot navigation capability of each method, we use Grounded-SAM to replace the vision modules of methods marked by an asteroid ($*$), which aligns with our model.}

\vskip 0.05in

  \centering
  \resizebox{.8\textwidth}{!}{%
  \begin{tabular}{lccccccc}
    
    \toprule
    \multirow{2}{*}{Method} & \multirow{2}{*}{Planner} &\multirow{2}{*}{Training-free} & \multirow{2}{*}{LLM} & \multicolumn{2}{c}{HM3D} &\multicolumn{2}{c}{HSSD}  \\
     \cmidrule(l{1pt}r{1pt}){5-6}\cmidrule(l{1pt}r{1pt}){7-8}
    &&&& Success$\uparrow$ & SPL$\uparrow$ & Success$\uparrow$ & SPL$\uparrow$ \\
    \midrule
    Random Exploration$^{*}$& Random & $\checkmark$ & - & 26.5 & 9.2 & 30.2 & 12.7  \\
    \midrule
     Frontier \cite{yamauchi1997frontier}$^{*}$ & \multirow{2}{*}{Topological} & $\checkmark$ & - & 33.7  & 15.3 & 36.0 & 17.7  \\
     Voronoi$^{*}$ & & $\checkmark$ & - & 38.7  & 23.3 &  40.3 & 22.2  \\
    \midrule
    L3MVN \cite{yu2023l3mvn}$^{*}$ & \multirow{4}{*}{Semantic} & $\checkmark$ &GPT-2& 35.2  & 16.5 & 38.4 & 19.4  \\
     Pixel-Nav \cite{cai2023bridging} && \texttimes &GPT-4 & 37.9  & 20.5 & - & -  \\
     ESC \cite{zhou2023esc} && $\checkmark$ & GPT-3.5  & 39.2 & 22.3 & - & -  \\
     \textbf{VoroNav (Ours)} && $\checkmark$ &GPT-3.5 & \textbf{42.0}  & \textbf{26.0} & \textbf{41.0} & \textbf{23.2}  \\

    \bottomrule

  \end{tabular}%
    }
\vspace{-10pt}
  \label{tab:1}
\end{table*}

In order to assess the navigation capability and exploration efficiency of VoroNav, we carry out extensive experiments on two representative datasets: HM3D and HSSD datasets.

\subsection{Baselines and Metrics}
\label{subsec:exp_baseline}
{\bf Datasets.} The HM3D dataset provides $20$ high-fidelity reconstructions of entire buildings and contains $2$K validation episodes for object navigation tasks. The HSSD dataset provides $40$ high-quality synthetic scenes and contains $1.2$K validation episodes for object navigation.

{\bf Metrics.} 
We adopt Success rate (Success) and Success weighted by Path Length (SPL) as the evaluation metrics \cite{anderson2018evaluation}, which are defined as follows:
\vspace{-5pt}
\begin{itemize}[leftmargin=10pt, itemsep=-2pt]
\item \textbf{Success} represents the percentage of successful episodes to the total number of episodes. 
\item \textbf{SPL} quantifies the agent's mobility efficiency in goal-oriented navigation by calculating the inverse ratio of the actual path length traversed to the optimal path length weighted by success rate.
\end{itemize}
\vspace{-5pt}
{\bf Baselines.} 
We conduct comparative evaluations of VoroNav and several representative baseline planners, including: 
\vspace{-5pt}
\begin{itemize}[leftmargin=10pt, itemsep=-2pt]
\item \textbf{Random Exploration} drives the robot to march to randomly sampled points in unexplored areas.
\item \textbf{Frontier} \cite{yamauchi1997frontier} is an exploration method that selects the nearest boundary points of unexplored areas and unoccupied areas as the mid-term goals.
\item \textbf{Voronoi} corresponds to VoroNav but without considering the semantic reward.
\item \textbf{L3MVN} \cite{yu2023l3mvn} is a region-oriented navigation method that leverages LLM to select the optimal mid-term waypoint by evaluating regions clustered by the frontier points.
\item \textbf{Pixel-Nav} \cite{cai2023bridging} is image-based zero-shot navigation that analyzes panoramic images and utilizes LLM to determine optimal pixel for exploration.
\item \textbf{ESC} \cite{zhou2023esc} is a pioneering method employing an LLM to determine the mid-term goal from the frontier points during exploration.
\end{itemize}
\vspace{-5pt}
Note that Random Exploration, Frontier, and Voronoi methods only utilize topological information of the map for planning, while L3MVN, Pixel-Nav, and ESC additionally require semantic information. 

\subsection{Results and Analysis}
\subsubsection{Comparison with SOTA Methods}
As shown in \Cref{tab:1}, our approach outperforms the best-performing competitor (+2.8\% Success and +3.7\% SPL on HM3D, +2.6\% Success and +3.8\% SPL on HSSD). 
As expected, the Random Exploration method suffers from the blind exploration strategy, resulting in a high likelihood of targeting the wrong areas and walking back and forth during exploration. 
The Frontier method consistently pursues the closest unexplored boundary, resulting in a more rapid exploration compared to the Random Exploration. 
Yet, the Frontier method suffers from relatively low efficiency as it fails to prioritize rapidly locating the target at the perception level. 
The Voronoi method shares similarities with the Frontier method in search for the nearest unexplored point.
However, Voronoi enhances the navigation process by proceeding to informative neighbor nodes along the RVG paths, thereby pursuing to uncover larger areas with very few steps. 
Both L3MVN and ESC adopt the frontier exploration strategy and leverage an LLM to select appropriate frontier points, whereas Pixel-Nav makes decisions after a fixed number of steps, utilizing LLMs to predict the direction with the highest probability of leading to the target, and subsequently employing an RGB-based policy to plan a route and navigate accordingly.
These three semantic planning methods uniformly make decisions at predetermined intervals, which can lead to agents determining the mid-term goal in suboptimal positions with insufficient observations, thereby failing to fully unleash the reasoning power of LLM. Our VoroNav method further improves the navigation process while using scene descriptions of broader observations to assist LLM in reasoning and decision-making, thereby achieving better performance. 

\begin{figure*}[t]
  \centering
   \includegraphics[width=0.8\linewidth]{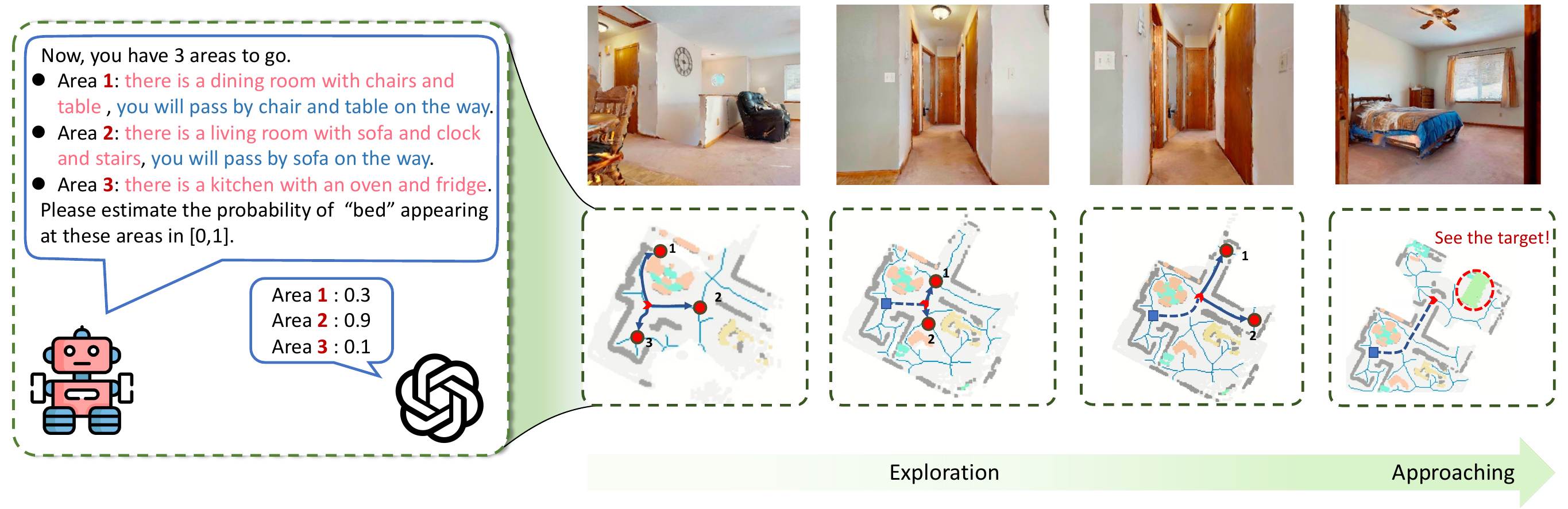}

   \caption{\textbf{Simulation Experiments.} Utilizing LLM, the agent explores efficiently, discovers the target with a minimal path cost, and finally navigates to the target object with success. In this figure, we visualize the RGB images and semantic maps of the four global decision instances, and the dialog box on the left exhibits the conversation between the agent and LLM in the first global decision process.}
   \label{fig:experiments}
   \vspace{-10pt}
\end{figure*}

\Cref{fig:experiments} illustrates a successful episode of VoroNav navigating to the target with the help of the RVG and LLM. It visualizes the observations and the environments at four key global decisions in this episode, and the details of a representative LLM decision-making process.

\subsubsection{Ablation Study}
To manifest the contribution of each module, we compare VoroNav with three ablation models on both HM3D and HSSD datasets. 
The Voronoi method keeps the same settings as in \Cref{subsec:exp_baseline}. 
The Voro-path method omits the farsight descriptions in VoroNav and depends entirely on the path descriptions, whereas the Voro-farsight method excludes the path descriptions in VoroNav and depends merely on the farsight descriptions for decision-making. 
As indicated in \Cref{tab:2}, both Voro-path and Voro-farsight show higher Success and SPL than Voronoi, indicating the benefits of integrating semantic information to augment navigation capabilities.
Furthermore, VoroNav exhibits superior performance compared to all ablation models, demonstrating the positive outcomes of integrating both path and farsight descriptions to enhance the performance of LLM's reasoning.

\begin{table}[t]
  \caption{\textbf{Ablation Study.} We compare VoroNav with three ablation models: (1) Voronoi: Voronoi-based navigation without any semantic guidance; (2) Voro-path: VoroNav that eliminates the farsight descriptions; (3) Voro-farsight: VoroNav that eliminates the path descriptions.}

\vskip 0.05in
  
  \centering
  \resizebox{0.45\textwidth}{!}{
  \begin{tabular}{ccccc}
    \toprule
    \multirow{2}{*}{Method}& \multicolumn{2}{c}{HM3D} &\multicolumn{2}{c}{HSSD}  \\
     \cmidrule(l{1pt}r{1pt}){2-3}\cmidrule(l{1pt}r{1pt}){4-5}
    & Success$\uparrow$ & SPL$\uparrow$ & Success$\uparrow$ & SPL$\uparrow$ \\
    \midrule
    Voronoi &38.7 & 23.3  & 40.3& 22.2  \\
    Voro-path  & 40.0 & 24.2& 40.6& 23.0 \\
    Voro-farsight & 41.2 & 25.2& 40.8& 22.7 \\
    \textbf{VoroNav}  & \textbf{42.0}  & \textbf{26.0} & \textbf{41.0} & \textbf{23.2}  \\

    \bottomrule
  \end{tabular}
}
  \label{tab:2}
  \vspace{-10pt}
\end{table}

\subsubsection{Planning Study}
To verify that Voronoi-based methods are more suitable for visual navigation compared to Frontier-based methods, we introduce two metrics to evaluate the planning effect of each method: the Success weighted by Collision Avoidance (SCA) and the Success weighted by Explored Area (SEA).

The metric SCA quantifies the proportion of non-collision steps to the total number of agent forward steps weighted by success rate, manifesting the tendency of obstacle avoidance in navigation, which is defined as follows:
\begin{equation}
  SCA=\frac{1}{N} \sum_{i=1}^{N} S_{i}  \left(1-\eta \frac{C_{i}}{F_{i}}\right),
  \label{eq:SCA}
\end{equation}
where $N$ is the total number of episodes, the binary variable $S_i\in\{0,1\}$ indicates whether the $i^\text{th}$ episode is successful ($S_i=1$) or not ($S_i=0$). 
The discount factor $\eta$ is predetermined and varies across datasets.
We set $\eta = 0.1$ in HM3D and $\eta = 1$ in HSSD.
The variables $C_i$ and $F_i$ represent the number of collisions and forward steps in the $i^\text{th}$ episode.

The metric SEA measures the normalized ratio of the observed region to the path length weighted by success rate, denoting the efficiency of perceiving the surrounding environment, which is defined as follows: \begin{equation}
  SEA=\frac{1}{N} \sum_{i=1}^{N} S_{i} \left(\gamma\frac{\sqrt{A_i}}{L_i}\right),
  \label{eq:SCE}
\end{equation}
where $A_i$ and $L_i$ signify the area of the explored region and path length in the $i^\text{th}$ episode, respectively. The discount factor $\gamma$ is also predetermined, and we set $\gamma = 0.002$ in HM3D and $\gamma = 0.01$ in HSSD.

As demonstrated in \Cref{tab:4}, we find that the SCA and SEA of Voronoi-based methods (Voronoi and VoroNav) significantly surpass those of Frontier-based methods (Frontier and L3MVN). 
The higher SCA score suggests that, throughout the exploration process, the mid-term goals of the Voronoi-based methods are typically chosen at intersections within unoccupied regions, which are less likely to be in proximity to obstacles, thereby reducing the incidence of collisions compared to the frontier-based methods. 
Specifically, if the agent gets too close to obstacles during navigation, it is easy to encounter blind spots in perceiving the surrounding environment, which can lead to collisions with undetected obstacles that lie within these blind spots during subsequent movement. 
In contrast, our method involves less movement near obstacles during exploration, allowing for wider unobstructed views and thus more comprehensive observations, resulting in safer navigation. 
Similarly, a higher SEA score indicates that the Voronoi-based methods favor intersections rich in information, enabling broader areas to be observed with minimal movement. 
The enhanced perceptual range increases the probability of the agent discovering the target directly. 
Meanwhile, this strategy yields potentially more valuable scene hints generated by comprehensive perception to inform LLM's decision-making, thus heightening the chances of locating the target object.

\begin{table}[t]
  \caption{\textbf{Planning Study.}
  We analyze the planning capability of different methods by comparing SCA and SEA. Our VoroNav method achieves the highest SCA and SEA scores, indicating advanced capability for obstacle avoidance and low-cost exploration.}

\vskip 0.05in
  
  \centering
  \resizebox{0.37\textwidth}{!}{%
  \begin{tabular}{ccccc}
    \toprule
    \multirow{2}{*}{Method}& \multicolumn{2}{c}{HM3D} &\multicolumn{2}{c}{HSSD}  \\
     \cmidrule(l{1pt}r{1pt}){2-3}\cmidrule(l{1pt}r{1pt}){4-5}
    & SCA$\uparrow$ & SEA$\uparrow$ & SCA$\uparrow$ & SEA$\uparrow$ \\
    \midrule
    Frontier &24.2&17.4&35.5&16.5  \\
    Voronoi  &29.4&17.9&40.2&18.6 \\
    L3MVN &27.5&17.7&37.6&16.6 \\
    \textbf{VoroNav}  &\textbf{39.8}&\textbf{20.9}&\textbf{40.9}&\textbf{19.3}  \\

    \bottomrule
  \end{tabular}%
    }
    \vspace{-20pt}
  \label{tab:4}
\end{table}

\section{Conclusion}
\label{sec:con}
We have presented the VoroNav framework that explores a novel form of graph representation for navigation space and substantially enhances ZSON by using a structured graph-based exploration strategy. 
Our approach circumvents the limitations of traditional end-to-end and map-based methods by generating informative waypoints and representing the environment with an innovative fusion of text information. 
The RVG generation module, together with the use of GPT-3.5 for decision-making, leads to more strategic navigation and efficient exploration. 
By making use of LLM and topologically structured scene graphs, VoroNav sets a new benchmark for ZSON and opens up new pathways for intelligent robotic systems to interact with environments.

\section*{Impact Statements}
VoroNav, in addressing the challenges of autonomous navigation in robotics, has profound implications for the future of household robotics and AI-driven navigation systems. By leveraging the synergy of semantic mapping and LLMs, it promises enhanced efficiency and effectiveness in robotic navigation tasks, potentially transforming how machines interact with and understand their environments. VoroNav enhances the navigation process with an explicit reasoning process during navigation, using a Reduced Voronoi Graph integrated with LLMs. 
Ethically, VoroNav represents a stride in responsible AI application, balancing technological advancement with the need for safe, reliable, and intuitive robot behavior in domestic settings. To address safety and ethical concerns, our experiments are conducted in controlled, simulated environments using open-source datasets. This ensures the predictability and safety of the agent's behaviors. However, the reliability of generative models in practical applications remains an area for future research. Key to this advancement is improving the LLMs' accuracy and precision in planning and sequential action prediction, crucial for ensuring safety in real-world deployment.

\bibliography{main}
\bibliographystyle{icml2024}

\newpage
\appendix
\twocolumn

\section{Method Details}

\subsection{RVG Generation}
\label{appendix_RVG}
\begin{figure}[htbp]
  \centering
   \includegraphics[width=0.8\linewidth]{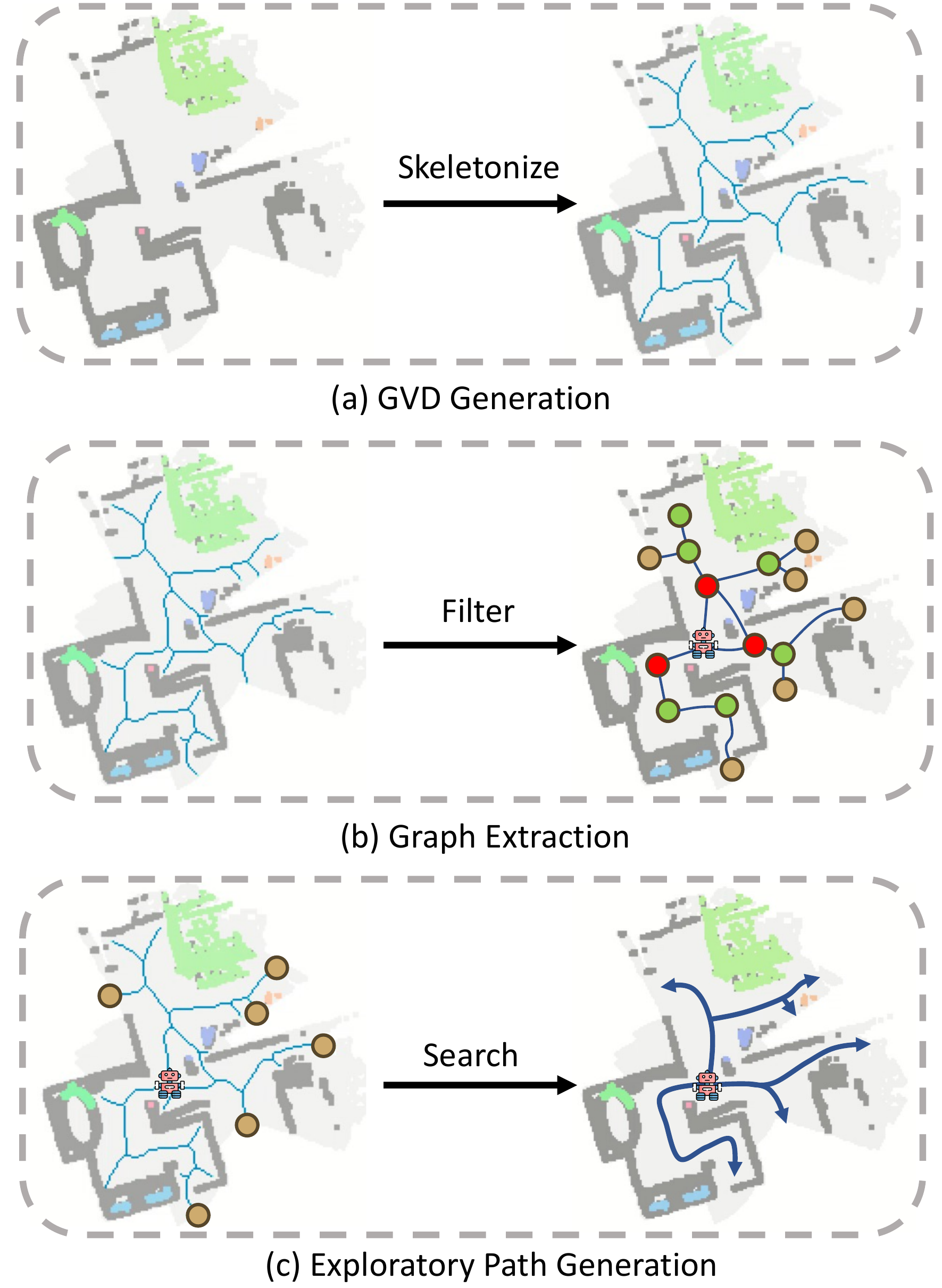}

   \caption{
   \textbf{The Voronoi Processing Module.} In (a), we skeletonize the areas that are not occupied by obstacles in the explored area and obtain the GVD (blue lines). In (b), the nodes and edges are extracted in GVD to form the RVG. The agent nodes (robot icon), neighbor nodes (red circles), ordinary nodes (green circles), and exploratory nodes (orange circles) are filtered by the location of the nodes. In (c), the exploratory paths (blue arrows) are generated by searching for the shortest paths on the GVD from the agent node to the exploratory nodes. 
   }
   \label{fig:voronoi}
\end{figure}

Given the obstacle and explored maps, we can obtain the unoccupied map by logically subtracting the obstacle map from the explored map, representing the traversable areas within the observed regions. 
We then preprocess the unoccupied map by using the morphology methods \cite{van2014scikit} to fill holes and smooth boundaries. 
To obtain the GVD, we extract a set of Voronoi points $\mathcal{V}$ by skeletonizing the unoccupied map (\Cref{fig:voronoi} (a)). 
Subsequently, to manifest the connectivity and accessibility of the unoccupied map, the GVD can be processed into RVG $\mathcal{G}$, a graph form with nodes $V$ and edges $E$ (\Cref{fig:voronoi} (b)). 
The nodes correspond to GVD points that are either at intersections or on the endpoint of GVD, while the segments directly connecting two adjacent nodes are identified as edges. 
The raw graph is then preprocessed through operations such as merging proximate nodes and eliminating trivial forks. 

\begin{figure}[H]
  \centering
   \includegraphics[width=\linewidth]{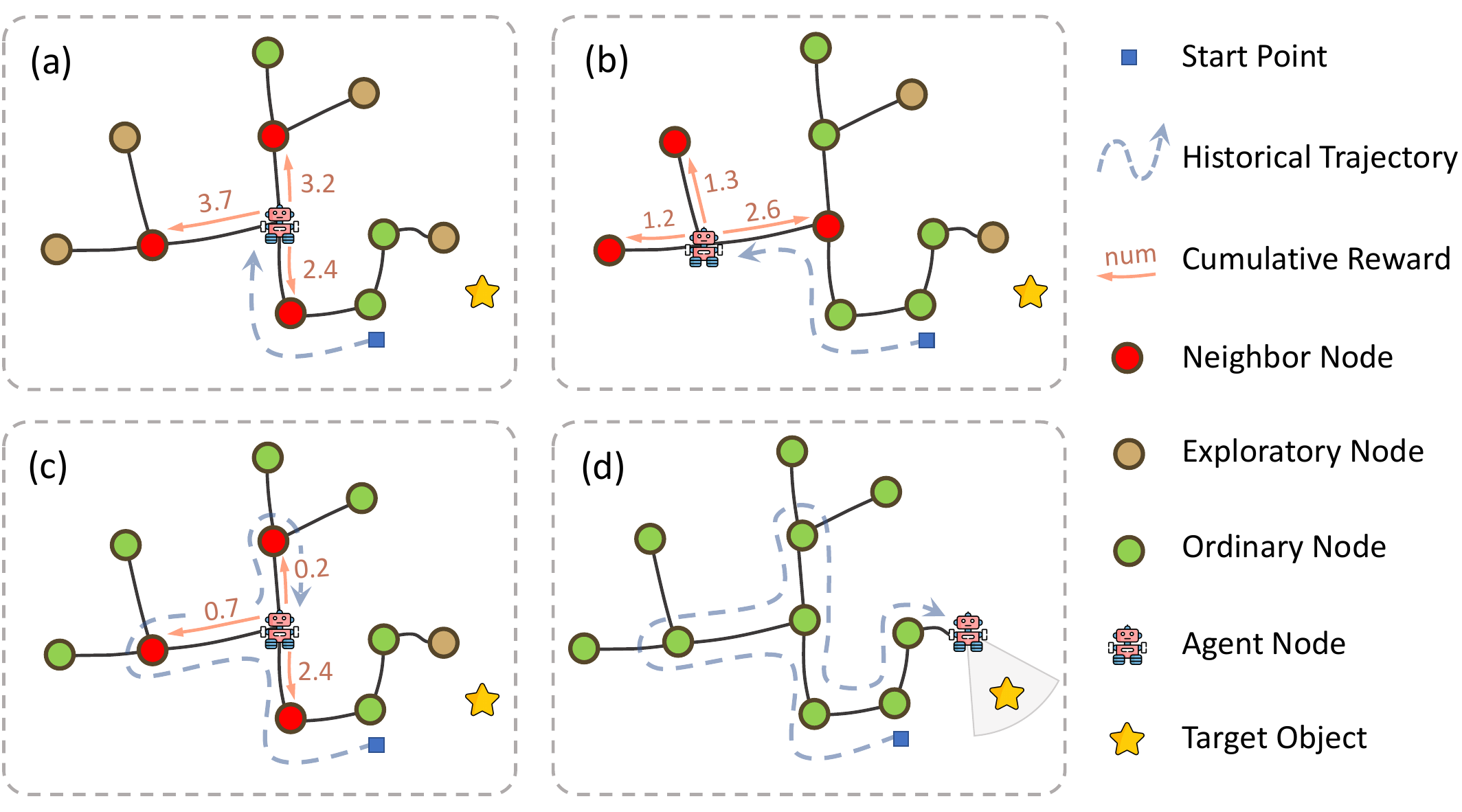}

   \caption{\textbf{A Worst-case Example.}
     This figure depicts the exploration process on RVG and the cumulative rewards of neighbor nodes when LLM alone makes undesirable decisions. (a) Three neighbor nodes are traversed by exploratory paths, one of which has been passed by the agent, and the agent compares LLM scores between the other two to make a choice. (b) Two nodes have not been passed by the agent but are not traversed by exploratory paths, and another node that the agent has passed by but with an exploratory path traversed is selected as a mid-term goal. (c) Three neighbor nodes have all been passed by the agent, but one of them leads to an unexplored area and is selected as a mid-term goal. (d) The agent finds the target object after exploration.
   }
   \label{fig:graph}
\end{figure}

\vspace{-0.5cm}
\subsection{Reward Roles}

To illustrate the robustness of VoroNav in various challenging scenarios and the roles played by distinct reward vectors, we have visualized a worst-case navigation example and the details of cumulative rewards, as shown in \Cref{fig:graph}. For exploration purposes, only neighbor nodes traversed along the exploratory path indicate its heading toward unexplored areas, which is foremost for the exploration process (refer to \Cref{fig:graph} (c)); Concerning efficiency, we rely on the agent's historical decisions, considering the current agent node as the optimal choice from previous decisions and discouraging turning back. If there is one or more extensions from the agent node leading to unexplored areas, the agent is inclined to continue its ongoing exploration (refer to \Cref{fig:graph} (a)). Conversely, if no extension offers exploratory paths, it indicates that unexplored regions exist elsewhere, and historically traversed nodes will be revisited, prompting the agent to return to previously traveled paths (refer to \Cref{fig:graph} (b)). Thus, the agent prioritizes exploration and efficiency from topological perspectives in the navigation. In cases where multiple nodes hold equivalent exploration and efficiency rewards, the agent will proceed to the neighbor node where the target object is more likely to be found, as indicated by the higher predicted semantic probability (refer to \Cref{fig:graph} (a)).

\subsection{Navigation Process}
\label{sec:nav_proc}
A complete process of a navigation episode is illustrated in \Cref{algo:nav}.
The code snippet of the \textit{LookAround} procedure, as presented in \Cref{algo:nav}, is further elaborated in \Cref{algo:lookaround}. 
At the beginning of each episode, the subgoal is initially empty. 
At each step, the agent updates the semantic map of its surroundings and the RVG accordingly (Line 4-8). 
If the agent detects the target object at any time, it will immediately plan a direct route to approach the object (Line 9-10). 
Conversely, if the target remains undetected, the agent performs a complete rotation to establish a preliminary RVG scene representation (Line 13, \Cref{algo:lookaround}). 
The agent then navigates to the closest RVG node (Line 14). 
Upon reaching the RVG node or the mid-term goal(Line 16), the agent rotates a full circle again (Line 17, \Cref{algo:lookaround}), derives the exploratory paths (Line 18) and surrounding images (Line 19), generates corresponding descriptions of paths (Line 20-21) and farsight (Line22) integrated with the respective neighbor nodes. 
A large language model is then employed to evaluate the fused descriptions of each neighbor node, obtaining semantic rewards based on the results of scene reasoning (Line 23). 
Concurrently, the agent acquires exploration rewards (Line 24) and efficiency rewards (Line 25) on the neighbor nodes by analyzing the layout of unexplored areas and historical trajectories. The neighbor node with the highest cumulative reward will be selected as the optimal mid-term goal point for exploration (Line 26). Finally, low-level motion planning is utilized to devise a sequence of actions targeting the mid-term goal (Line 29). If the agent reaches the vicinity of the identified target or takes actions surpassing the maximum number of steps, it will immediately issue a ``Stop'' action and the episode ends.

\begin{algorithm}[ht]
   \caption{Navigation Process of VoroNav}
   \label{algo:nav}
\begin{algorithmic}[1]
   \STATE {\bfseries Input:} Target object $G$
   \STATE {\bfseries Initialize:} Initial observation $\mathcal{O}_0 \leftarrow \varnothing$\\
    \hspace*{1.57cm}Initial semantic map $\mathcal{M}_0 \leftarrow \varnothing$\\
    \hspace*{1.57cm}Step Number $t \leftarrow 1$\\
    \hspace*{1.52cm} $\textit{SubGoal} \leftarrow \textit{None}$\\    
   \WHILE{Episode is done}
      \STATE $\mathcal{O}_t \leftarrow \mathcal{O}_{t-1} \bigcup \{p_t, I_t\}$
      \STATE \textit{ObjectMasks} $\leftarrow$ GroundedSAM$(I_t)$
      \STATE $\mathcal{M}_t \leftarrow $ Mapping$(\mathcal{M}_{t-1}, \mathcal{O}_t, \textit{ObjectMasks})$
      \STATE $\mathcal{V} \leftarrow \text{Skeletonize}(\mathcal{M}_t)$
      \STATE $\mathcal{G} \leftarrow (V, E) \leftarrow \text{Filter}(\mathcal{M}_t, \mathcal{V})$
      \IF{$G$ exists in $\mathcal{M}_t$}
         \STATE $\textit{SubGoal} \leftarrow \text{Location}(\mathcal{M}_t,G)$
      \ELSE
         \IF{SubGoal is None}
            \STATE LookAround;
            \STATE $\textit{SubGoal} \leftarrow \text{Nearest}(p_t, V)$
         \ENDIF
         \IF{Agent reaches node in $V$}
            \STATE LookAround
            \STATE $\mathcal{P} \leftarrow \text{Search}(p_t, \mathcal{G})$
            \STATE $\textit{NeighborImages} \leftarrow \text{Select}(\mathcal{I}_t, \mathcal{M}_t, V)$
            \STATE $\textit{PathPrompt} \leftarrow \text{Template}(\mathcal{M}_t, \mathcal{G}, \mathcal{P})$
            \STATE $\mathcal{D}^p \leftarrow \text{LLM}(\textit{PathPrompt})$
            \STATE $\mathcal{D}^f \leftarrow \text{BLIP}(\textit{NeighborImages})$
            \STATE $\mathbf{L} \leftarrow \text{LLM}(G, \mathcal{D}^p, \mathcal{D}^f)$
            \STATE $\mathbf{P} \leftarrow \text{Exploration}(V, E)$
            \STATE $\mathbf{C} \leftarrow \text{Efficiency}(\mathcal{M}_t, \mathcal{O}_t, V)$
            \STATE $\textit{SubGoal} \leftarrow \text{Decision}(V, \mathbf{P}, \mathbf{C}, \mathbf{L})$
         \ENDIF
      \ENDIF
      \STATE $a_t \leftarrow \text{FMM}(\mathcal{M}_t, p_t, \textit{SubGoal})$
      \STATE $t \leftarrow t+1$
   \ENDWHILE
   \STATE {\bfseries Result:} Episode ends.
\end{algorithmic}
\end{algorithm}

\section{Experiment Details}

\subsection{Experiment Setup} 
Evaluations on HM3D follow settings outlined in the Habitat ObjectNav challenge 2022 \cite{habitatchallenge2022}. 
Evaluations on HSSD adopt the same validation parameters utilized in \cite{khanna2023habitat}. 
In both datasets, the agent is a LoCoBot \cite{gupta2018robot} with a base radius of 0.18m, outfitted with an RGB-D camera mounted at a height of 0.88 meters and a pose sensor that provides accurate localization. 
The camera has a $79^\circ$ \textit{Horizontal Field of View} (HFoV) and frame dimensions of $480\times640$ pixels. 
The agent's action space is \{Stop, MoveForward, TurnLeft, TurnRight, LookUp, LookDown\}, with a discrete movement increment of $0.25m$ and discrete rotations of $30^\circ$. 
The object goal categories in episodes include ``bed'', ``chair'', ``sofa'', ``tv'', ``plant'', and ``toilet''.

\begin{algorithm}[H]
\caption{LookAround}
\label{algo:lookaround}
\begin{algorithmic}[1]
\STATE $\mathcal{I}_{t+12} \leftarrow \varnothing$
\FOR{$i \leftarrow 1$ {\bfseries to} $12$}
    \STATE $a_t \leftarrow \textit{TurnRight}$
    \STATE $t \leftarrow t+1$
    \STATE $\mathcal{O}_t \leftarrow \mathcal{O}_{t-1} \bigcup \{p_t, I_t\}$
    \STATE \textit{ObjectMasks} $\leftarrow$ GroundedSAM$(I_t)$
    \STATE $\mathcal{M}_t \leftarrow$ Mapping$(\mathcal{M}_{t-1}, \mathcal{O}_t, \textit{ObjectMasks})$
    \STATE $ \mathcal{V} \leftarrow \text{Skeletonize}(\mathcal{M}_t)$
    \STATE $\mathcal{G} \leftarrow (V, E) \leftarrow \text{Filter}(\mathcal{M}_t, \mathcal{V})$
    \STATE $\mathcal{I}_{t+12-i} \leftarrow \mathcal{I}_{t+12-i} \bigcup \{I_t\}$
\ENDFOR
\IF{$G$ exists in $\mathcal{M}_t$}
    \STATE $\textit{SubGoal} \leftarrow \text{Location}(\mathcal{M}_t,G)$
    \STATE \textbf{continue}
\ENDIF

\end{algorithmic}
\end{algorithm}

\begin{table}[t]
  \caption{\textbf{Ground-truth Semantics.} We find that after replacing the Grounded-SAM with ground truth semantic segmentation, our VoroNav still achieves the best performance, revealing the superiority of our planning module.}

\vskip 0.05in
  
  \centering
  \begin{tabular}{ccccc}
    \toprule
    \multirow{2}{*}{Method}& \multicolumn{2}{c}{HM3D} &\multicolumn{2}{c}{HSSD}  \\
     \cmidrule(l{1pt}r{1pt}){2-3}\cmidrule(l{1pt}r{1pt}){4-5}
    & Success$\uparrow$ & SPL$\uparrow$ & Success$\uparrow$ & SPL$\uparrow$ \\
    \midrule
    Frontier &63.5&33.0&51.2&20.6  \\
    Voronoi  &67.3&37.4&57.2&32.7 \\
    L3MVN &65.5&36.5&58.0&28.0 \\
    \textbf{VoroNav}  &\textbf{67.6}&\textbf{40.5}&\textbf{59.7}&\textbf{34.1}  \\

    \bottomrule
  \end{tabular}

  \label{tab:3}
\end{table}

\begin{figure}[ht]
  \centering
   \includegraphics[width=\linewidth]{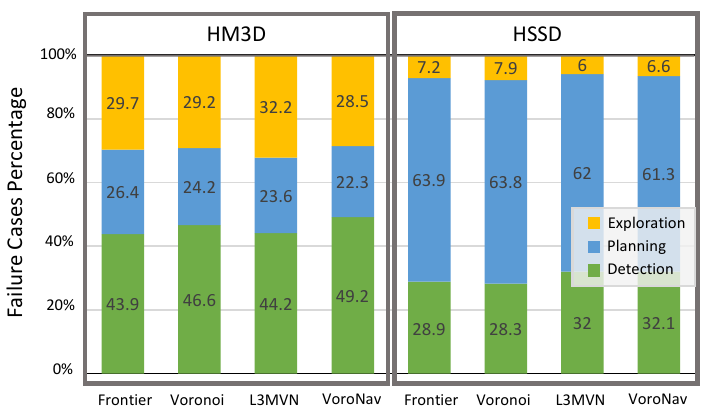}

   \caption{\textbf{Failure Cases Percentage.} In the failure statistics on HM3D and HSSD datasets, our VoroNav experiences the fewest planning failures.}
   \label{fig:failure}
\end{figure}

\subsection{Failure Case Study}
We have collected the reasons and frequencies of failures across all episodes, categorizing them into detection failures, planning failures, and exploration failures. 
Detection failure occurs when the agent mistakenly identifies non-target items as targets or overlooks the actual targets within the agent's field of view. 
Planning failure arises when the agent gets stuck or fails to navigate to the target location despite having accurately detected the target. 
Exploration failure is attributed to the situation where the agent has not encountered the target within the assigned maximum number of steps.
As shown in \Cref{fig:failure}, we find that most failures are caused by incorrect detection in HM3D and by abnormal planning in HSSD.

\subsection{Ground-truth Vision Experiments}
To eliminate the impact of detection errors during navigation and analyze the planning and exploration effects of navigation methods, we uniformly replace the RGB images across all methods with ground-truth semantic images. 
As shown in \Cref{tab:3}, when semantic priors of reasoning and planning are absent, the Voronoi method exhibits considerably superior performance in terms of both Success and SPL compared to the Frontier method. 
This enhancement in performance proves that within the topological domain, the Voronoi planning approach is more suitable for goal-oriented navigation than the Frontier exploration method. 
With the guidance of LLM, VoroNav outperforms L3MVN by a large margin, further demonstrating that VoroNav's information processing and decision-making are more effective for robot navigation with lower step cost and a higher success rate compared to L3MVN.

\onecolumn



\subsection{Prompt Template}

\begin{figure*}[h]
  \centering
   \includegraphics[width=0.7\linewidth]{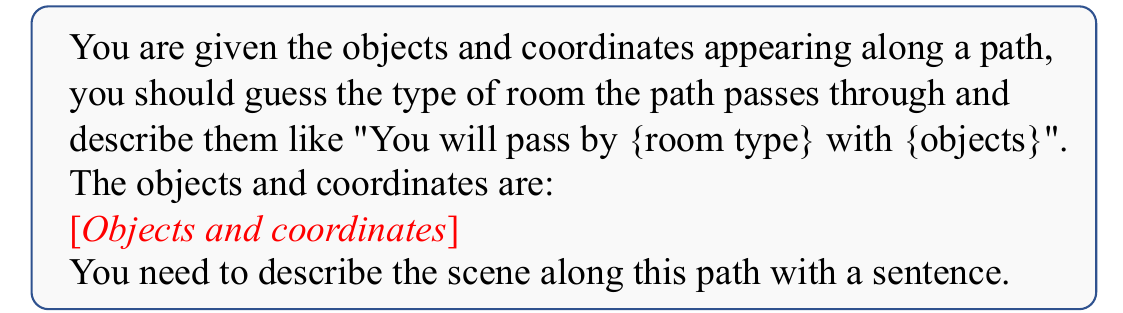}

   \caption{Prompt Template for Generating Path Description.}
   \label{fig:path_description_prompt}
\end{figure*}

\begin{figure*}[h]
  \centering
   \includegraphics[width=0.7\linewidth]{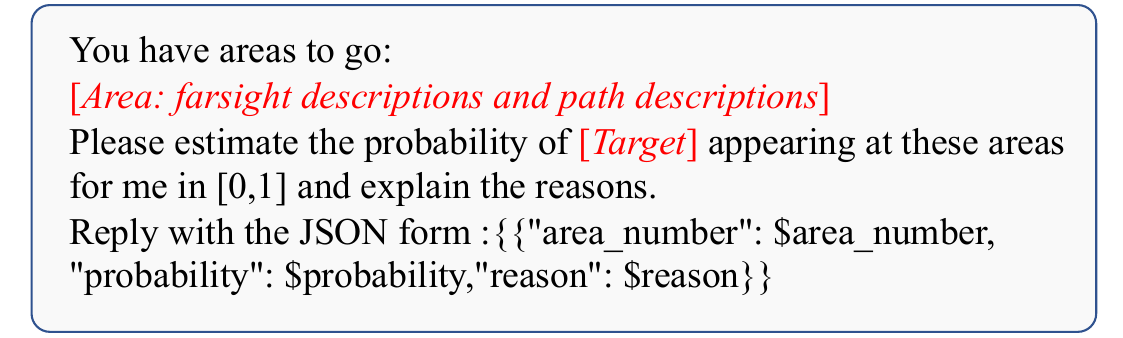}

   \caption{Prompt Template for LLM Decision.}
   \label{fig:decision_prompt}
\end{figure*}

In \Cref{fig:path_description_prompt} and \Cref{fig:decision_prompt}, we present the specific prompt templates for path description generation and decision-making. The red font in the prompt templates refers to the parts that vary according to different scenarios. 

\subsection{Examples of LLM Conversation}
\begin{figure*}[h]
  \centering
   \includegraphics[width=\linewidth]{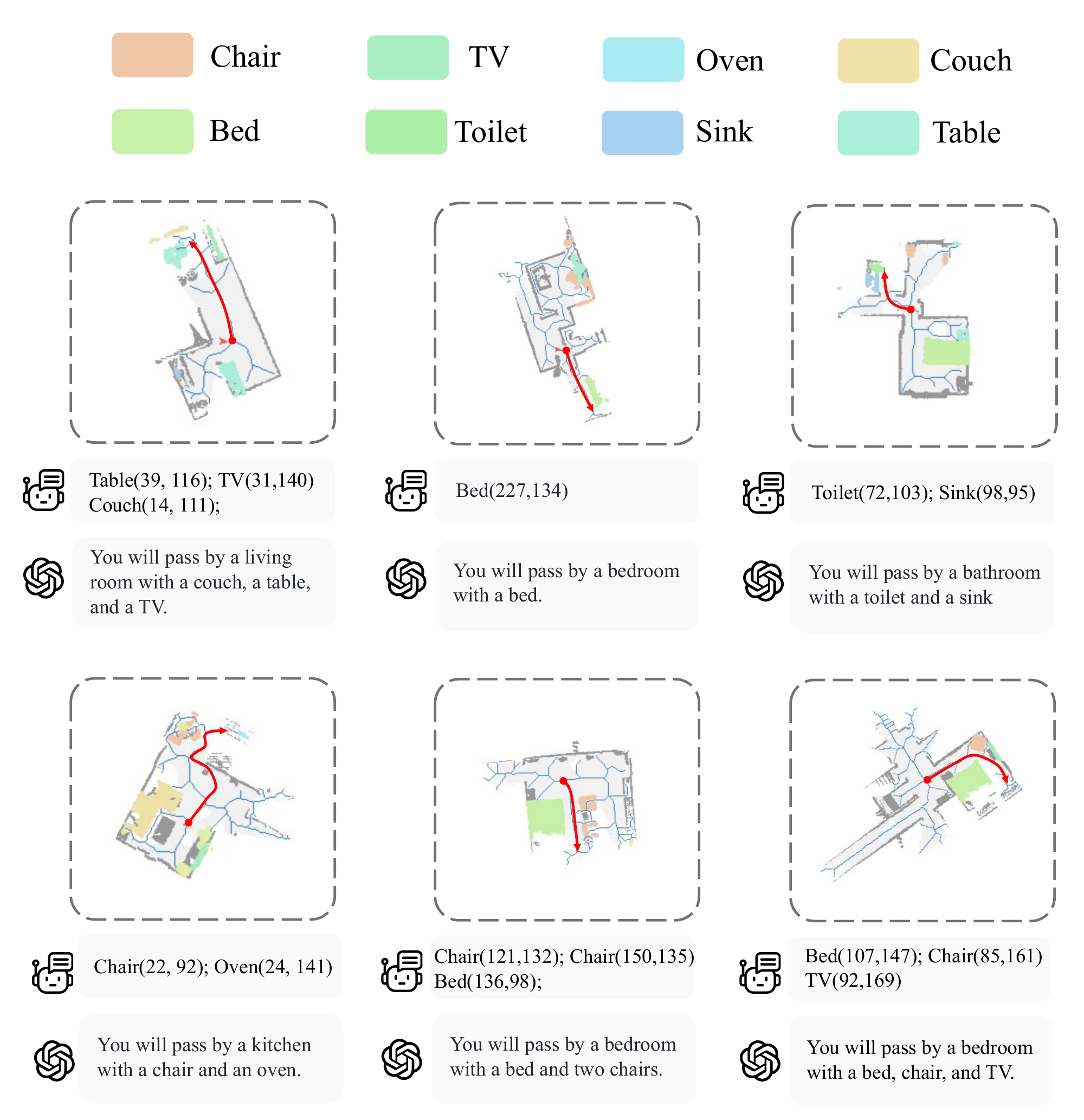}

   \caption{Examples of Path Description Generation.}
   \label{fig:path_example}
\end{figure*}

\begin{figure*}[h]
  \centering
   \includegraphics[width=\linewidth]{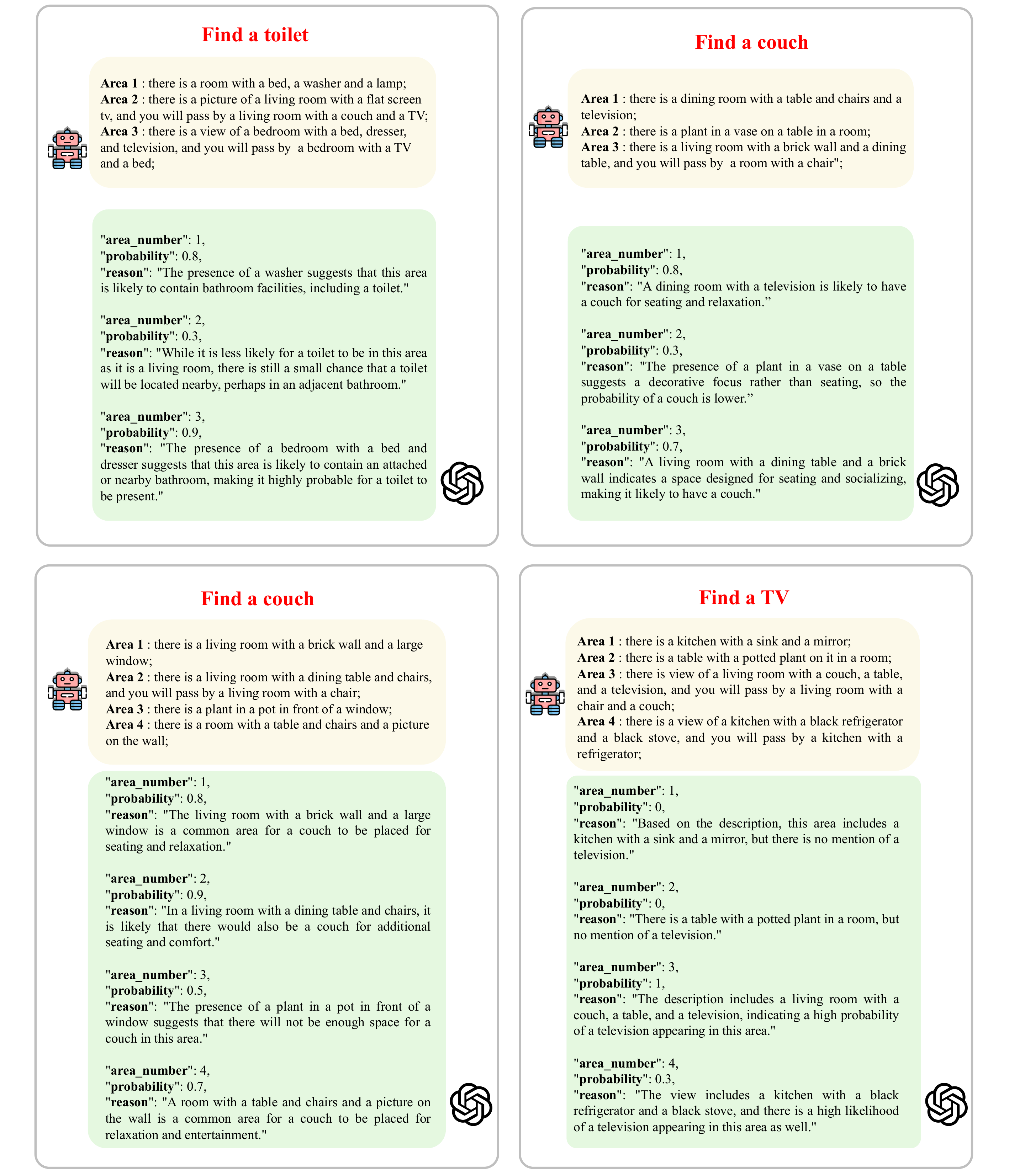}

   \caption{Examples of LLM Decision.}
   \label{fig:decision_example1}
\end{figure*}

\begin{figure*}[h]
  \centering
   \includegraphics[width=\linewidth]{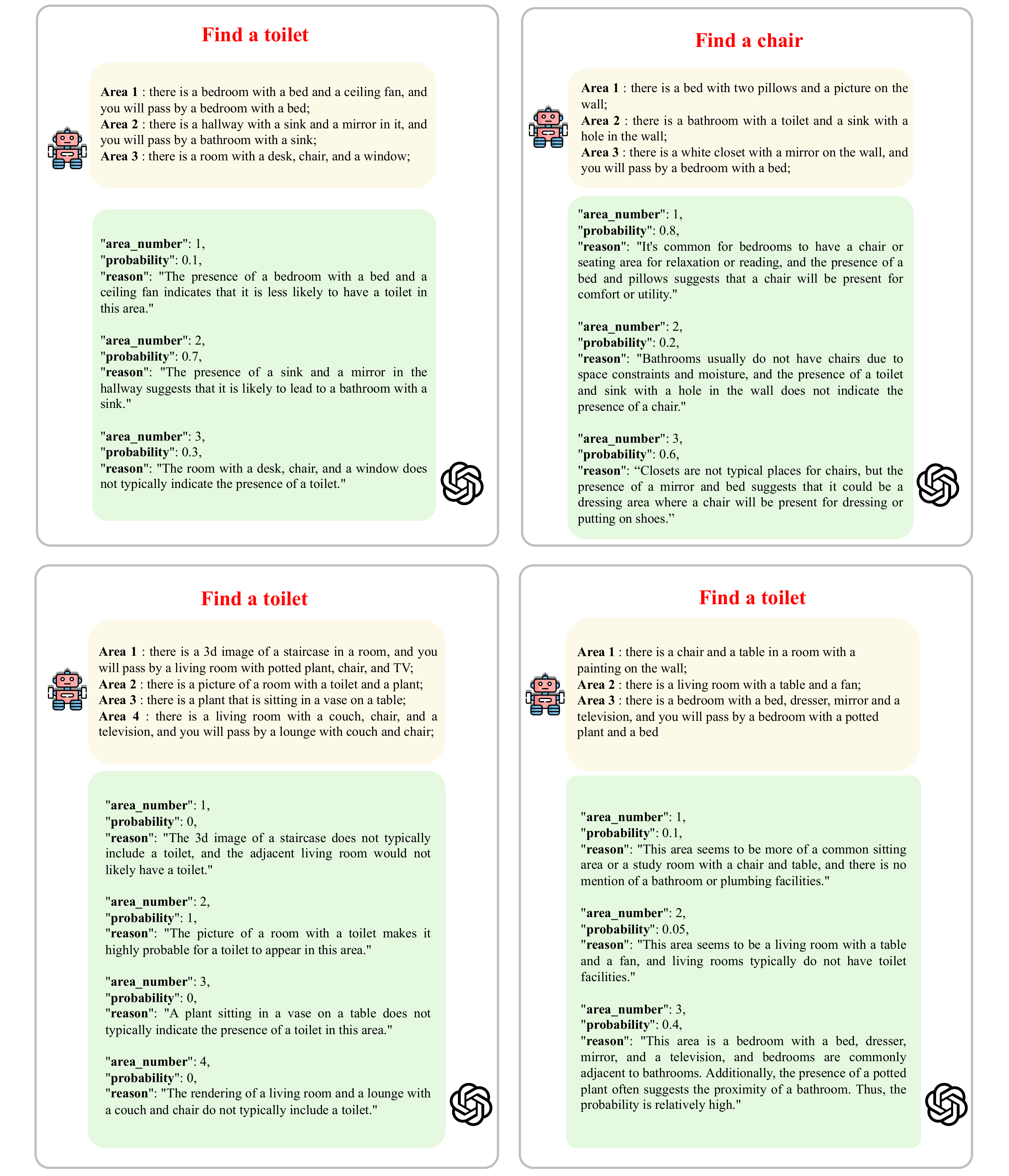}

   \caption{Examples of LLM Decision.}
   \label{fig:decision_example2}
\end{figure*}

We show supplemental examples of path description generation in \Cref{fig:path_example}. LLM decision examples are shown in \Cref{fig:decision_example1} and \Cref{fig:decision_example2}.

\end{document}